\documentclass{isprs}
\usepackage{setspace}
\usepackage{geometry} 
\usepackage{epstopdf}
\usepackage[labelsep=period]{caption}
\usepackage{siunitx}
\usepackage{etoolbox}
\robustify\bfseries
\usepackage{upgreek}
\usepackage[hidelinks]{hyperref}
\usepackage{tabularx}
\usepackage{booktabs}
\usepackage{subfig}
\usepackage{multirow}
\usepackage{color}
\usepackage{natbib}
\usepackage{paralist}
\usepackage{amsmath}
\usepackage[super]{nth}
\usepackage[nameinlink]{cleveref}
\usepackage{float}
\usepackage{stfloats}

\begin{document}

\title{Developing a machine learning framework for estimating soil moisture with VNIR hyperspectral data}
\author{
S. Keller\textsuperscript{1}, F. M. Riese\textsuperscript{1}, J. St\"otzer. \textsuperscript{1}, P. M. Maier\textsuperscript{1}, S. Hinz\textsuperscript{1}}
\address{
	\textsuperscript{1 }Institute of Photogrammetry and Remote Sensing, Karlsruhe Institute of Technology, Karlsruhe, Germany 
	\\(sina.keller, felix.riese, johanna.stoetzer, philipp.maier, stefan.hinz)@kit.edu\\
}

\commission{I, }{VI}
\workinggroup{I/1}
\icwg{}

\abstract{In this paper, we investigate the potential of estimating the soil-moisture content based on VNIR hyperspectral data combined with LWIR data. 
Measurements from a multi-sensor field campaign represent the benchmark dataset which contains measured hyperspectral, LWIR, and soil-moisture data conducted on grassland site.
We introduce a regression framework with three steps consisting of feature selection, preprocessing, and well-chosen regression models.
The latter are mainly supervised machine learning models.
An exception are the self-organizing maps which combine unsupervised and supervised learning. 
We analyze the impact of the distinct preprocessing methods on the regression results.
Of all regression models, the extremely randomized trees model without preprocessing provides the best estimation performance.
Our results reveal the potential of the respective regression framework combined with the VNIR hyperspectral data to estimate soil moisture measured under real-world conditions.
In conclusion, the results of this paper provide a basis for further improvements in different research directions.
} 
\maketitle

\section{INTRODUCTION}
\label{sec:introduction}

\sloppy

Precise data about spatial distributions and dynamics of soil moisture is valuable in many scopes of environmental applications.
Hydrological as well as meteorological processes are influenced by soil moisture. 
Besides soils, microbes, and plants depend heavily on it~\citep{farrelly2011sitka,cavagnaro2016soil,tian2018soil}.
Apart from this, soil moisture emerges as one of the key variables relating to hydrological disasters such as flash floods on a catchment scale~\citep{gill2006soil}. 
Soil-moisture data of e.g. catchments areas functions as input variable for estimating and mitigating flood impacts to enhance flood models~\citep{massari2014potential}.
In many regions, the soil-moisture distribution varies during the season of a year.
Thus, e.g. summer soil-moisture availability serves as a relative indicator of a potential rate of fire spread, fire intensity, and fuel consumption~\citep{girardin2009summer}. 
In addition, measured soil-moisture data is used to model the germination of seedbeds after such wildfires~\citep{flerchinger2004modelling}. 
Other studies have been conducted which refer to the linkages between soil moisture and wind erosion~\citep{wang2014freezethaw}.   
All these fields of application have in common that they require soil-moisture estimations under almost real-world conditions such as a soil surface covered with vegetation.

The demand for spatial coverage and temporal resolution of soil moisture varies widely in the fields of application.
Small-scale measurements, e.g. field site scale (pedon-scale), are performed with handheld sensors in combination with point-wise in situ soil-moisture measurements.
One advantage of this scale is a high temporal resolution. 
Large-scale observations rely on airborne and satellite-based remote sensing solutions~\citep[cf.][]{maggioni2006multi,john1992soil, finn2011remote,colini2014hyperspectral}.
Therefore, they cover catchments and larger areas with a limited temporal resolution.
Hence, a gap between the spatial coverage and temporal soil-moisture resolution as well as spatial coverages occurs~\citep{robinson2008soil}.

Developments in hyperspectral remote sensing during the last four decades have enhanced the data acquisition regarding e.g. spectral resolution for evaluating the soil-moisture dynamics.
Terrestrial hyperspectral remote sensing sensors mounted on drones can cover a pedon-scale and are able to retrieve spectral signatures of the soil-moisture distribution in-between the top-soil layers~\citep{kaleita2005relationship}.
The surface of such sites is characterized by inhomogeneous covers including different vegetation, soil, and  rock.
This inhomogeneity of the soil surface results in overlaying reflectance spectra and poses a challenge to identify the soil-moisture state and dynamic~\citep{salisbury1992emissivity}.
Since the datasets dealing with soil-moisture content are conducted expensively in field campaigns or laboratory measurements, most of them are of limited size.

When it comes to the estimation or modeling of soil-moisture contents based on remote sensing data, two trends can be deduced generally. 
First,  hyperspectral sensors, which combine a fine spatial resolution and narrow bandwidths, outperform multispectral-retrieved data, especially in heterogeneous areas.
Second, the short-wave infrared (SWIR) hyperspectral sensors obtain better results in estimating soil-moisture contents than the visible and near infrared (VNIR) sensors~\citep{dalal1986simultaneous,finn2011remote}.
Crucial disadvantages of the SWIR sensors are the high acquisition cost, the need for active cooling, and, as direct consequence, the large weight and complex handling when mounting on e.g. a drone.
Referring to these barriers and despite the knowledge of the great potential of such SWIR sensors, we seek to address the estimation of soil-moisture dynamics based on hyperspectral data in the wavelength of 450 nm to 950 nm (VNIR\footnote{We refer to this range of wavelength as \textit{visible and near infrared} (VNIR) range due to reasons of simplicity.}).
Furthermore, long-wavelength infrared (LWIR) data measured with an thermal camera is used.

For our investigations, we chose a dataset which has been measured in a multi-sensor field campaign on a pedon-scale with defined surface conditions and precise monitoring of the soil-moisture dynamics. 
Real-world conditions, such as a vegetation cover, and therefore the ability to transfer applied methods are sustained.  
To provide a first impression, ~\citet{keller2018modeling} have described the multi-sensor field campaign.
The underlying pedo-hydrological processes monitored by several sensors as well as preliminary estimations of soil-moisture values are presented.
In contrast to this, the present contribution exemplifies the potential of the frequently underrated VNIR with respect to the subsurface soil-moisture retrieval. 
We evaluate a multitude of machine learning models which are suitable to solve non-linear regression problems with high-dimensional input data.
Furthermore, we investigate the ability of the machine learning framework to link the measured VNIR reflectance data of a vegetated soil surface to the measured subsurface soil-moisture data without additional domain-knowledge like spectral information of vegetation.

The main contributions of this paper are:
\begin{compactitem}
	\item a detailed investigation of the potential of VNIR hyperspectral data combined with LWIR data to estimate subsurface soil moisture;
	\item an appropriate regression framework based on ten regression models such as partial least square (PLS), an artificial neural network (ANN), and a self-organizing map (SOM) framework which merges unsupervised and supervised learning;
	\item a comprehensive evaluation of the regression performance and an analysis of the potential of the underlying sensor data for the estimation of subsurface soil-moisture dynamics on a field site scale in regards to hydrological application.
\end{compactitem}

We give a short overview on related work in regards to estimating soil moisture based on hyperspectral data with and without machine learning methods in \Cref{sec:rw}. 
Subsequently, we describe the measured dataset used for the evaluation of the several machine learning models of the framework.
The presentation of the methods follows in \Cref{sec:methods}. 
In \Cref{sec:results}, we evaluate the proposed machine learning models. 
Finally, we conclude our studies in \Cref{sec:conclusion}, respond to the overlying regression problem and give an overview about future applications of the pedon-scale soil-moisture estimation based on hyperspectral data.

\section{Related Work}
\label{sec:rw}

Traditionally, soil-moisture as well as pedo-hydrological dynamics and states are monitored with point-based in situ measurements using e.g. time domain reflectometry (TDR) probes and tensiometers.
Temporally high-resolution data can be aggregate based on these sensors.
The advantages of these techniques are the precise measurement of the vertical soil-moisture distribution at specific point locations.
However, to obtain area-wide insight, the traditional efforts are limited, time-consuming, and, depending on the experimental setup, uncertain~\citep{jackisch2017form}.

At this point, the employment of hyperspectral remote sensing techniques, covering the visible and near-infrared (VNIR), near-infrared (NIR), short-wave infrared (SWIR), and the LWIR range comes into effect.
The performance to estimate soil moisture based on VNIR, NIR and SWIR data enhances with increasing wavelengths~\citep{finn2011remote}.
The data acquisition with hyperspectral sensors ranges from point measurements with spectroradiometers to snapshots recordings by (drone-compatible) sensors or satellites.
The former provides a high spectral resolution, the latter advantages area-wide recordings.
Referring to~\citet{haubrock2008surface}, only few studies investigating surface soil moisture via airborne or spaceborne platforms record optical reflectance data.

Two distinct approaches are explored in regards to the estimation of soil-moisture contents especially with hyperspectral data.
The first approach focusses on engineering features by combing specific spectral bands to perform a ratio-calculation~\citep{vereecken2014on, fabre2015estimation, oltracarri2015improvement}.
The second approach relies on data-driven machine learning models which develop their potential when handling non-linear regression problems or processing large datasets like in case of satellite-based hyperspectral data~\citep{guanter2015the}. 
Most machine learning models are based on supervised learning such as partial least square (PLS) regression, random forest (RF), support vector machine (SVM), or artificial neural networks (ANN).
In addition,~\citet{riese2018introducing} introduce a framework of self-organizing maps for the regression of soil moisture which combines unsupervised and supervised learning.

According to the results of the feature engineering approaches (first approach), the SWIR spectrum includes the most important wavelengths which respond to soil-moisture contents~\citep{dalal1986simultaneous, wang2007a, haubrock2008surface, finn2011remote}.
A detailed review of modeling biomass and soil moisture with several remote sensing data and inter alia with machine learning is stated in~\citet{ifarraguerri2000unsupervised}. 
Further remote sensing data such as C-band polarimetric SAR or microwave scanning radiometry is also applied to estimate soil moisture in combination with machine learning~\citep{baghdadi2012estimation, pasolli2014polarimetric, xie2014soil}. 
These datasets are primarily conducted from satellite or airborne missions.

Generally, hyperspectral sensors provide spectral knowledge of surface conditions. 
The soil surface represents a key factor for the partitioning and redistributing of any precipitation before infiltrating into the subsurface~\citep{jarvis2007a, brooks2015hydrological}. 
Subsurface soil-moisture dynamics and states are estimated based on this spectral surface data combined with appropriate machine learning models.
Obviously, the spectral surface data represents an indirect approximation of the underlying physical soil-moisture processes.
Therefore, arising approximation uncertainties add to the yet existing model uncertainties.  
In sum, the benefits of hyperspectral applications prevail.

\section{Sensors and Dataset}
\label{sec:dataset}

\begin{figure*}[tb]
	\centering
	\subfloat[]{\includegraphics[height=5cm, keepaspectratio]{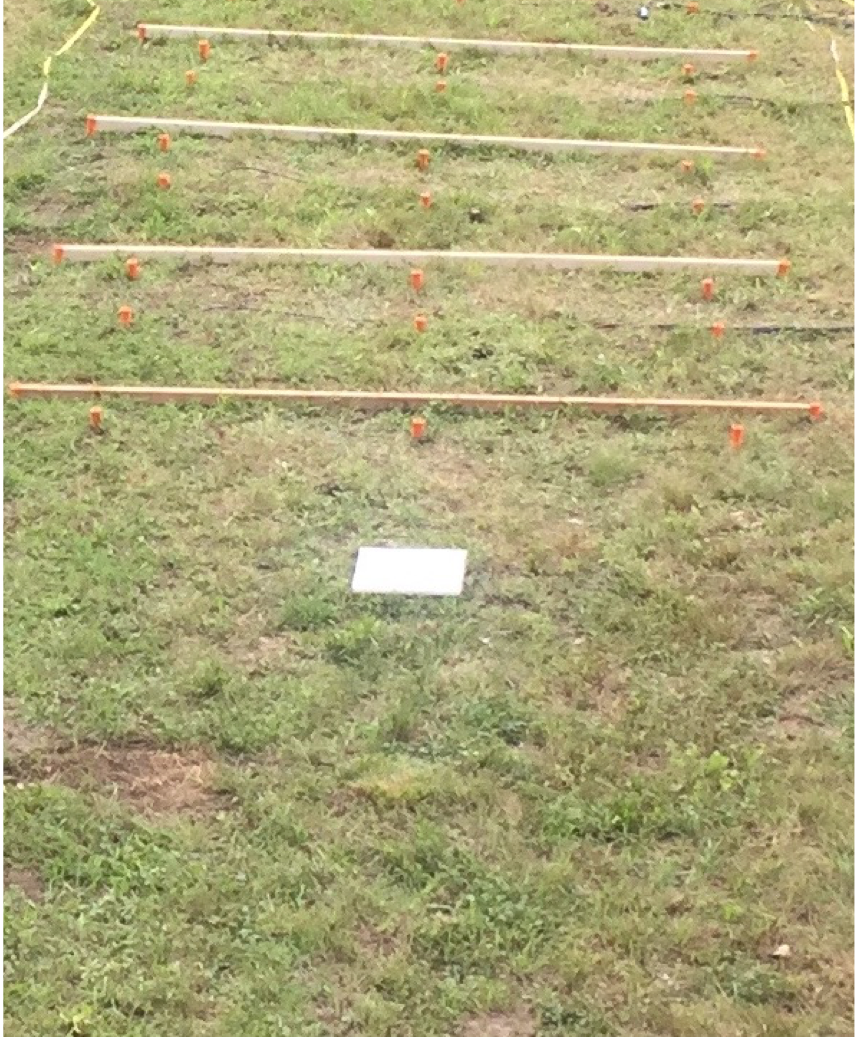}\label{fig:cam_rgb}}\quad
	\subfloat[]{\includegraphics[height=5cm, keepaspectratio]{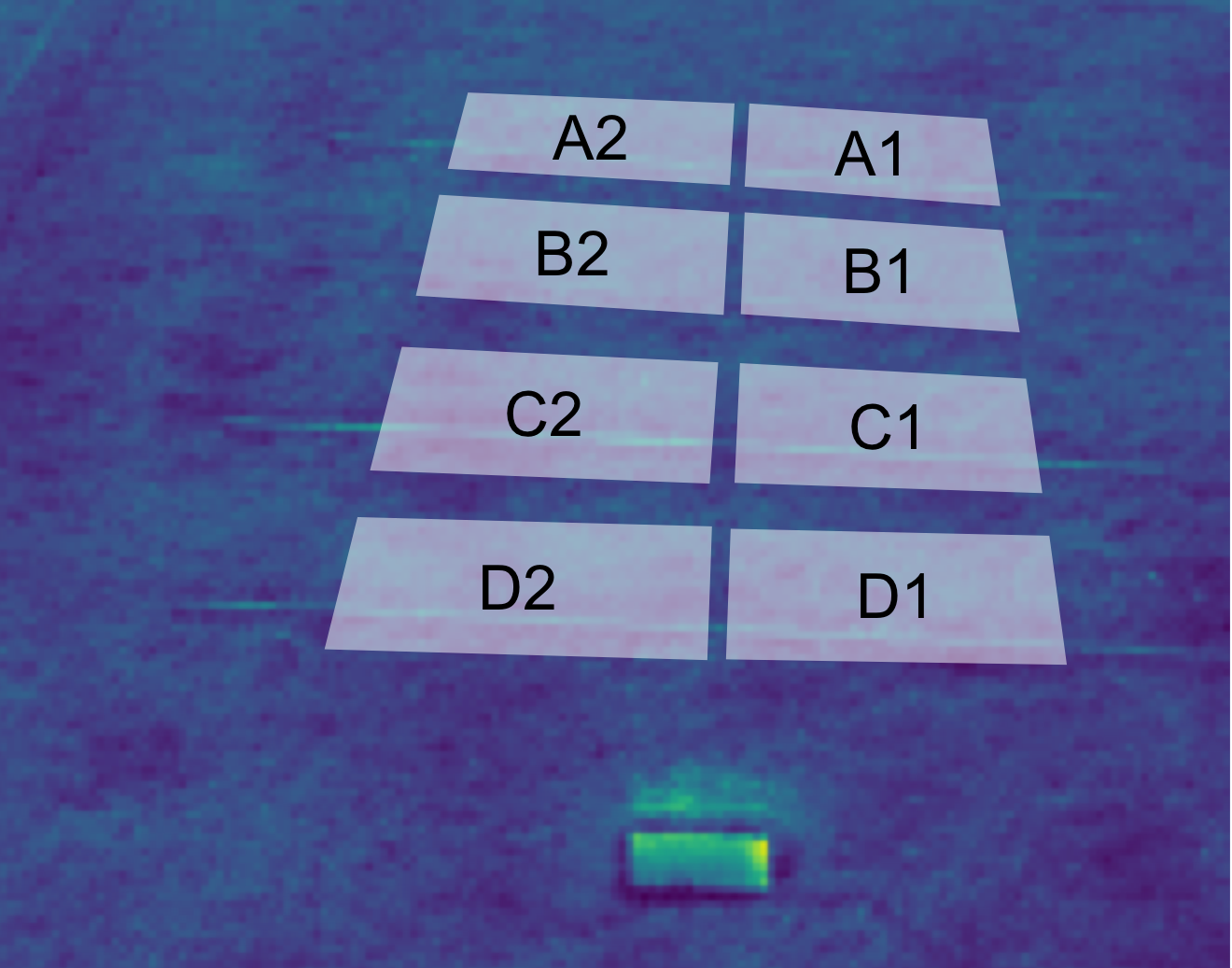}\label{fig:cam_hyp}}\quad
	\subfloat[]{\includegraphics[height=5cm, keepaspectratio]{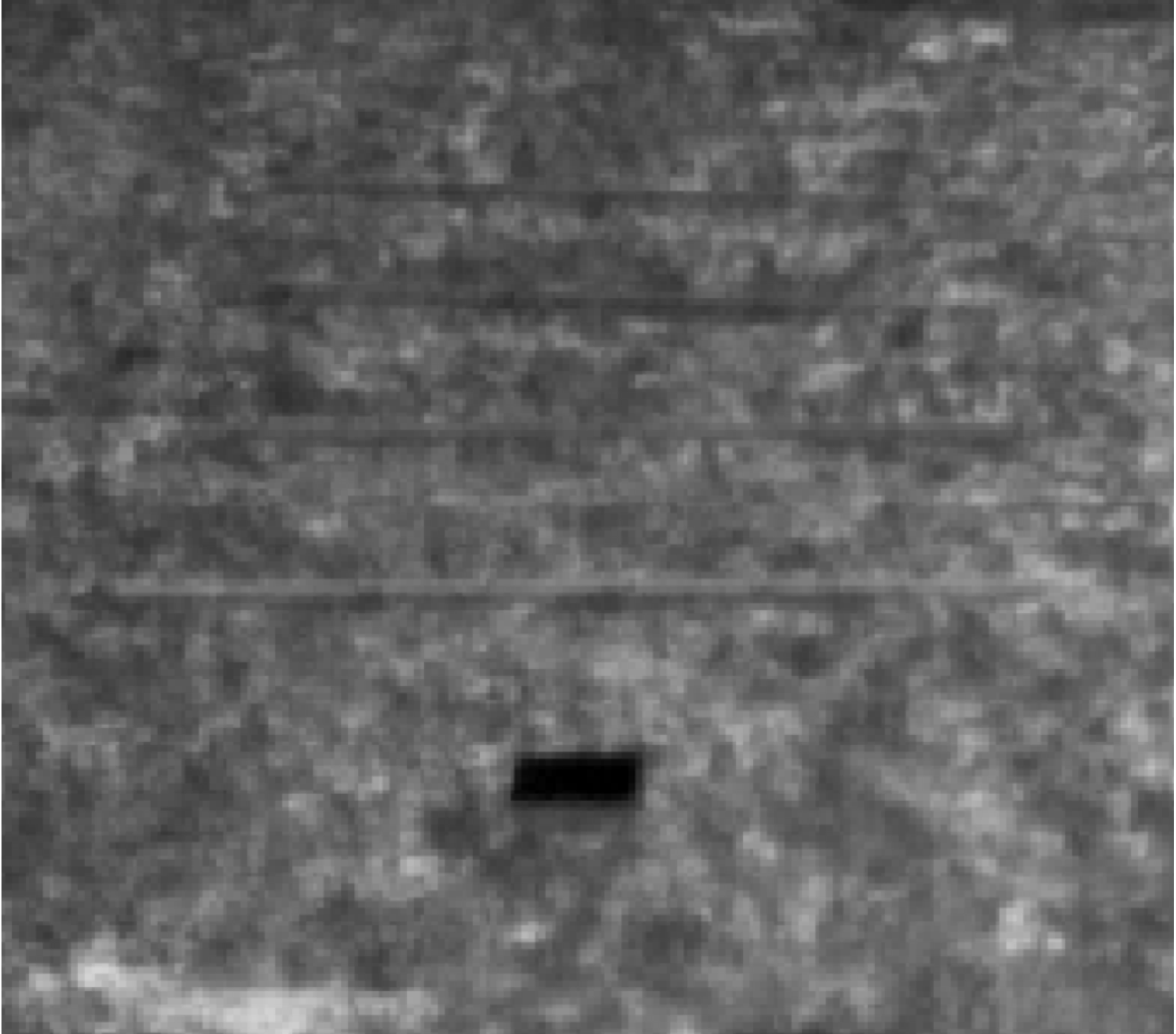}\label{fig:cam_ir}}
	\caption{An example of (a)~an RGB image, (b)~a hyperspectral snapshot, (c)~an LWIR image in false colors. The hyperspectral snapshot of $50\times 50$ pixels is pan-sharped to $1000\times 1000$ only to improve visualization.\label{fig:cam_all}}
\end{figure*}

To evaluate the potential of VNIR hyperspectral sensors as input data for the estimation of soil moisture, we rely on a  dataset which was conducted during a multi-sensor field campaign in August 2017 in Linkenheim-Hochstetten, Germany.
In this pedon-scale field campaign, the vegetated surface as well as soil-moisture states and dynamics have been monitored precisely. 
Since real-world conditions are sustained, the ability to transfer the applied regression methods is ensured.
A detailed overview of the field campaign with respect to the measurement setup as well as its constraints and the analysis of the pedo-hydrological processes can be found in~\citet{keller2018modeling}.
Eight plots of an undisturbed grassland site on loamy sand are the centerpiece of the campaign.
\Cref{fig:cam_rgb} shows the plot setup.
Each of these plots covers an area of one square meter and is irrigated according to a defined schema of various pulses~\cite[cf.][]{keller2018modeling}.  
Multiple time domain reflectometry (TDR) probes measure the soil moisture in various depths from \SI{2.5}{\centi\meter} to \SI{20}{\centi\meter}.  
Based on these sensors, pedo-hydrological states and dynamics after the irrigation processes are surveyed.
We refer to the TDR sensors in \SI{5}{\centi\meter} depth as soil-moisture reference and ground truth within the scope of the paper. 

A Cubert\footnote{Cubert GmbH, Ulm, Germany} UHD 285 hyperspectral snapshot sensor records the hyperspectral image data (cf. \Cref{fig:cam_hyp}).
The measured reflectance includes the spectral signatures of among others the soil surface covered with vegetation. 
The hyperspectral sensor is installed on a stage at \SI{10} {\meter} distance to cover the entire test site with one snapshot.
Each hyperspectral snapshot contains $50\times 50$ pixels and $125$ spectral channels ranging from \SIrange{450}{950}{\nano\meter} with a spectral resolution of \SI{4}{\nano\meter}.
The pan-sharpened $1000\times 1000$ pixels image in \Cref{fig:cam_hyp} only serves as improved visualization of the measurement area, we use the raw hyperspectral image for the regression framework.
As shown in \Cref{fig:cam_hyp}, the measurement angles differ between the eight plots in the field of view of the hyperspectral sensor due to the necessary setup of the whole field campaign~\cite[cf.][]{keller2018modeling}.   
All reflectance spectra in every image are normalized based on a white reference resulting in reflectance values between 0 and 1. 
The spectralon as white reference is positioned visually in each snapshot to ensure this normalization after the recording.
A thermal camera without active cooling (FLIR~\footnote{FLIR Systems. Inc., Portland, USA} Tau 2 640) records the LWIR images (cf. \Cref{fig:cam_ir}) and is installed next to the hyperspectral camera.
The LWIR images consist of $640\times 512$ pixels, each characterized by a temperature value in \si{\celsius}. 
With respect to the approximated position of the TDR probes in the subsurface, average spectra of each plot and recording are calculated for both remote sensing data.

\section{Methodology}\label{sec:methods}

\begin{figure}[tb]
	\centering
	\includegraphics[width=0.48\textwidth]{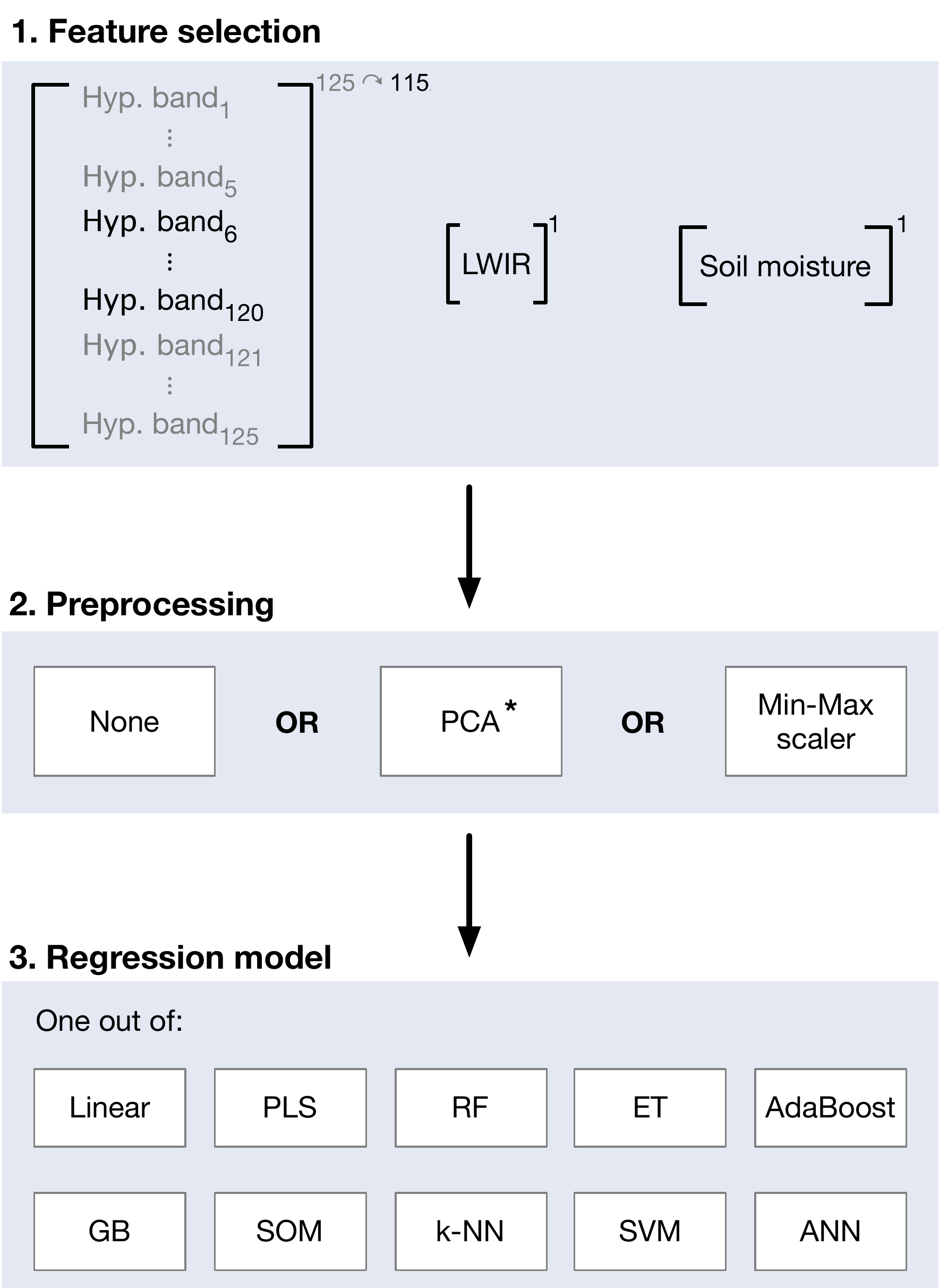}
	\caption{Schematic representation of the regression framework. \textsuperscript{*} The PCA is applied only on the hyperspectral and LWIR data. \label{fig:regschema}}
\end{figure}

Our proposed regression framework consists of three steps: the feature selection, the preprocessing, and the regression model to estimate soil moisture. 
\Cref{fig:regschema} represents the schema of the regression framework.

\subsection{Feature selection}
\label{sec:methods:sub:inputdata}

The regression is performed with the hyperspectral and LWIR image data as input vector and the soil-moisture data as target value.
The complete dataset consists of 1332 high-dimensional datapoints.
One datapoint is defined by $115$ selected hyperspectral bands, one LWIR value as well as one soil-moisture value as ground truth (cf. \Cref{fig:regschema}, top).
Five bands at the beginning and five bands at the end of the original $125$ hyperspectral bands are dismissed to avoid occurring sensor artifacts.

\begin{figure}[t]
	\centering
	\includegraphics[width=0.4\textwidth]{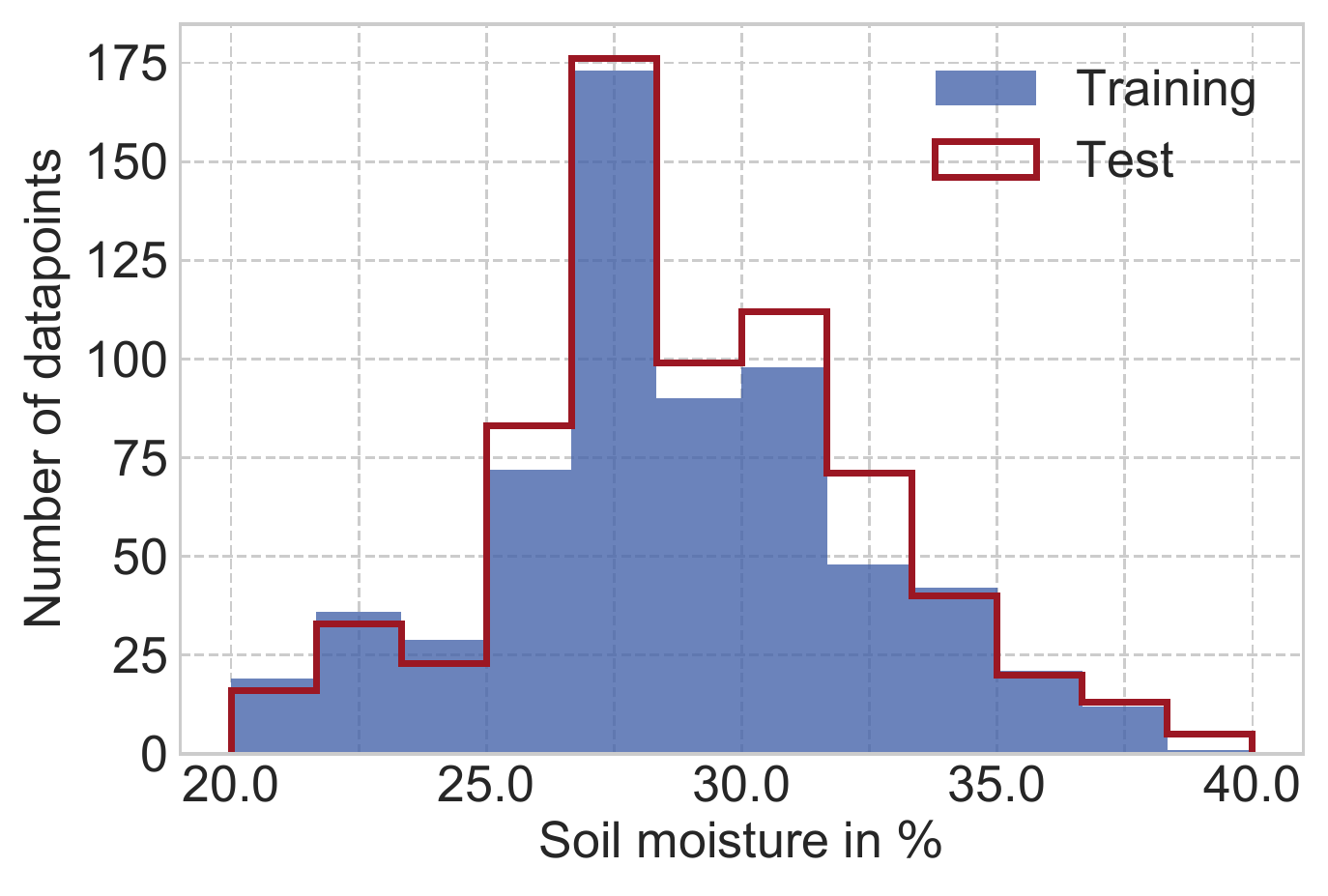}
	\caption{Distribution of the regression target variable (soil moisture) in the training (blue) and the test (red line) dataset.\label{fig:histtraintest}}
\end{figure}

For the regression framework, the complete dataset is split randomly into a training subset and a test subset. 
The training subset includes $641$ full datapoints, the test subset consists of $691$ full datapoints.
\Cref{fig:histtraintest} shows similar distributions of the measured soil-moisture values for the training and the test subsets.
This similarity enables a modeling of continuous soil-moisture values. 

\begin{figure}[t]
	\centering
	\includegraphics[width=0.45\textwidth]{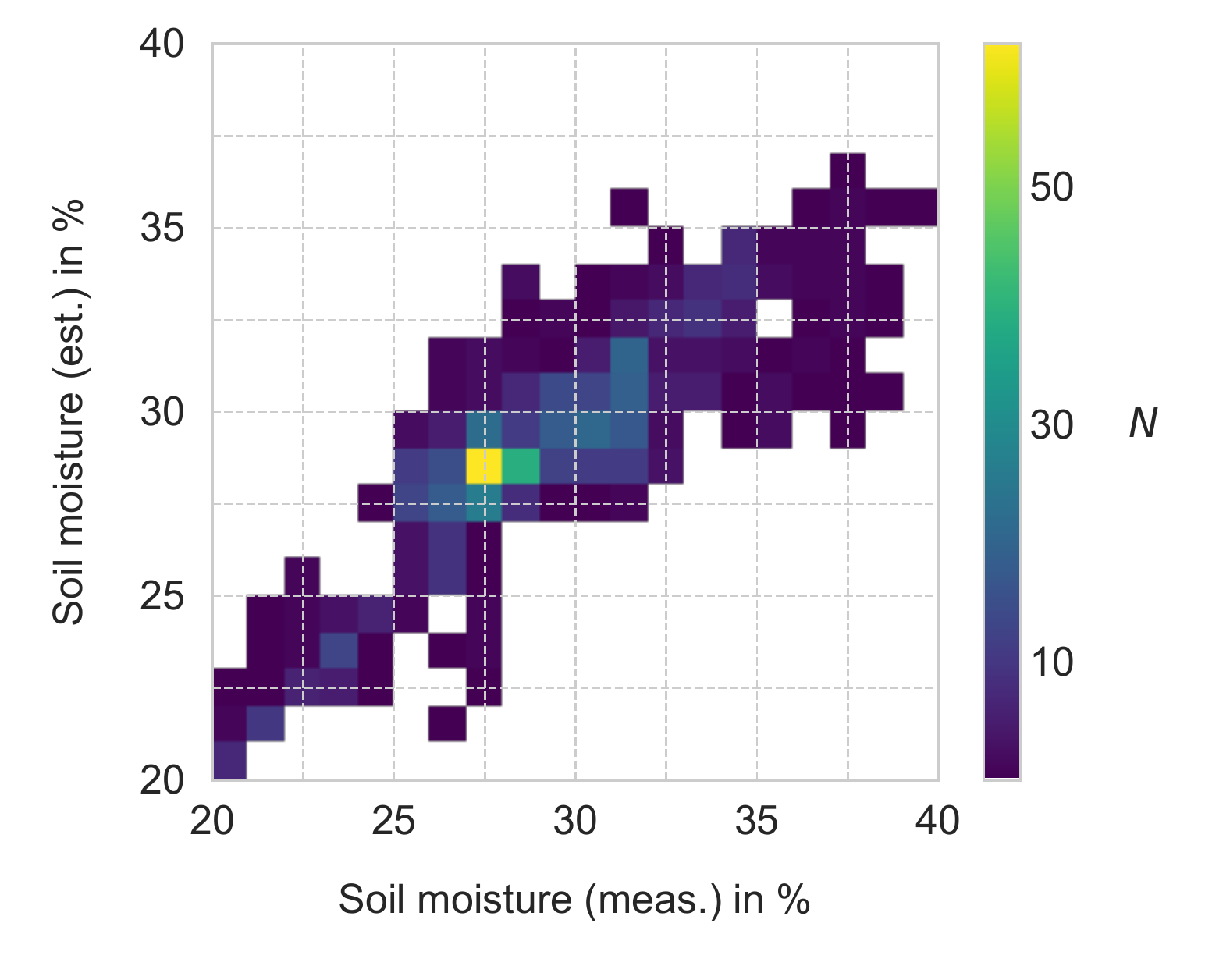}
	\caption{Example of a 2-dimensional histogram of the ET regressor showing the estimated vs. the measured soil-moisture values. $N$ represents the number of datapoints in the bin. \label{fig:2dhist}}
\end{figure}

\subsection{Preprocessing}
\label{sec:methods:sub:pre}

Aiming to estimate soil-moisture values based on hyperspectral and one-dimensional LWIR input data, we foster the regression by applying two distinct preprocessing methods.
The first method is a Principle Component Analysis (PCA) to reduce the dimensionality of the hyperspectral and LWIR input data (cf. \Cref{fig:regschema}).
It is applied to a stack of the VNIR reflectance values and the LWIR value. 
We use the first  \num{20} principal components for the regression, since they cover most of the dataset variances.
As second method, we apply a min-max scaling (cf. \Cref{fig:regschema}).
The scaling normalizes the input data to a fixed range between $0$ and $1$.
In contrast to the PCA, the min-max scaling uses all input data, including the soil-moisture values.

During the preprocessing step, we pick either the PCA for dimensionality reduction, the min-max scaling for normalization purposes, or no preprocessing (cf. \Cref{fig:regschema}, \nth{2} step, left) is performed.
We refer to the regression without preprocessing as the baseline framework.
Later, in the test phase of the regression model, the results of the estimation based on each preprocessing method is compared against the baseline prediction result without any preprocessing.

\subsection{Regression models}
\label{sec:methods:sub:reg}

To estimate soil moisture, we select appropriate regression models and include them to the framework (cf. \Cref{fig:regschema}, \nth{3}~step).
These are linear regression (least-squares), partial least squares (PLS), random forest (RF), extremely randomized trees (ET), adaptive boosting (AdaBoost), gradient boosting (GB), k-nearest-neighbors (k-NN), support vector machines (SVM), artificial neural networks (ANN), and a framework of self-organizing maps (SOM).
References as well as the implementations of these models are listed in \Cref{tab:hyperparams}.

During the training phase, the regression models are trained on the training subset by linking the hyperspectral and LWIR data to the soil-moisture target values.
Except for the SOM, all regressors perform the training phase exclusively supervised.
The SOM model includes two self-organizing maps to solve the regression problem, combining an unsupervised SOM with a supervised SOM. 
\citet{riese2018introducing} introduce the schema of this SOM model.

The parameters of a regression model are divided into hyperparameters and model parameters.
Model parameters are adapted during the training phase while hyperparameters are chosen beforehand.
The optimal setup of the hyperparameters changes depending on the preprocessing methods in step 2 of the regression framework.
\Cref{tab:hyperparams} shows exemplarily the setup of the hyperparameters for the baseline framework (no preprocessing in step 2).
We obtain a basic grid search with 10-fold cross validation on the training subset for each preprocessing method and regression model.

During the subsequent test phase, the trained regression framework estimates soil moisture on the basis of the hyperspectral and LWIR data of the test subset.
The estimated soil-moisture values (model predictions) are compared to the measured soil-moisture values.
The coefficient of determination $R^2$ and the root mean squared error (RMSE) express the regression performance.
Since the framework in general relies on randomization, we obtain all regression results by seven independent training procedures each with different random seeds.
The ensemble models RF, ET, AdaBoost, and GB provide additional information regarding the importance of the input variables (feature importance).

\section{Results and discussion}
\label{sec:results}

By applying PCA-based dimensionality reduction, we obtain regression results relying on the first 20 principle components.
By applying the min-max scaling to normalize the input data, the regression models rely on features in the range of $0$ to $1$.
Using ten regression models with supervised or the combination of unsupervised and supervised learning principles for estimating soil moisture, the respective results are depicted in \Cref{tab:results}.

\renewcommand{\arraystretch}{1.0}
\sisetup{detect-weight=true,detect-inline-weight=math}
\begin{table*}[p]
	\centering
	\caption{Regression results for the soil-moisture estimation.}
	\begin{tabular}{lSSSSSS}
		\toprule
		\multirow{ 3}{*}{Model} &\multicolumn{2}{c}{baseline} &\multicolumn{2}{c}{with PCA} &\multicolumn{2}{c}{with scaling}\\
		\cmidrule(ll){2-3}
		\cmidrule(ll){4-5}
		\cmidrule(ll){6-7}
		 & {$R^2$} & {RMSE} & {$R^2$} & {RMSE} & {$R^2$} & {RMSE}\\
		 & {in $\%$} &  {in $\%$ soil moisture} & {in $\%$} &  {in $\%$ soil moisture} & {in $\%$} & {in 1} \\
		\midrule
        Linear     & 50.7 & 2.5 & 49.3 & 2.6 & 50.7 & 0.1\\
		PLS        & 52.0 & 2.5 & 49.3 & 2.6 & 48.3 & 0.1\\
		RF         & 67.0 & 2.1 & 63.2 & 2.2 & 66.9 & 0.1\\
		ET         & \bfseries 73.0 & \bfseries 1.9 & \bfseries 69.1 & \bfseries 2.0 & \bfseries 72.8 & \bfseries 0.1\\
		AdaBoost   & 59.6 & 2.3 & 55.2 & 2.4 & 56.4 & 0.1\\
		GB         & 65.2 & 2.1 & 58.8 & 2.3 & 65.3 & 0.1\\
		k-NN       & 53.5 & 2.5 & 53.8 & 2.5 & 72.5 & 0.1\\
		SVM        & 50.7 & 2.5 & 50.2 & 2.6 & 70.4 & 0.1\\
		ANN        & 32.9 & 2.9 & 52.3 & 2.5 & 60.1 & 0.1\\
		SOM        & 42.5 & 2.7 & 43.0 & 2.7 & 56.5 & 0.1\\
		\bottomrule
	\end{tabular}
	\label{tab:results}
\end{table*}

Both linear regression models (linear and PLS regressors) perform the worst.
They are incapable of adapting to the high-dimensional regression problem.

\begin{figure}[tb]
	\centering
	\includegraphics[width=0.48\textwidth]{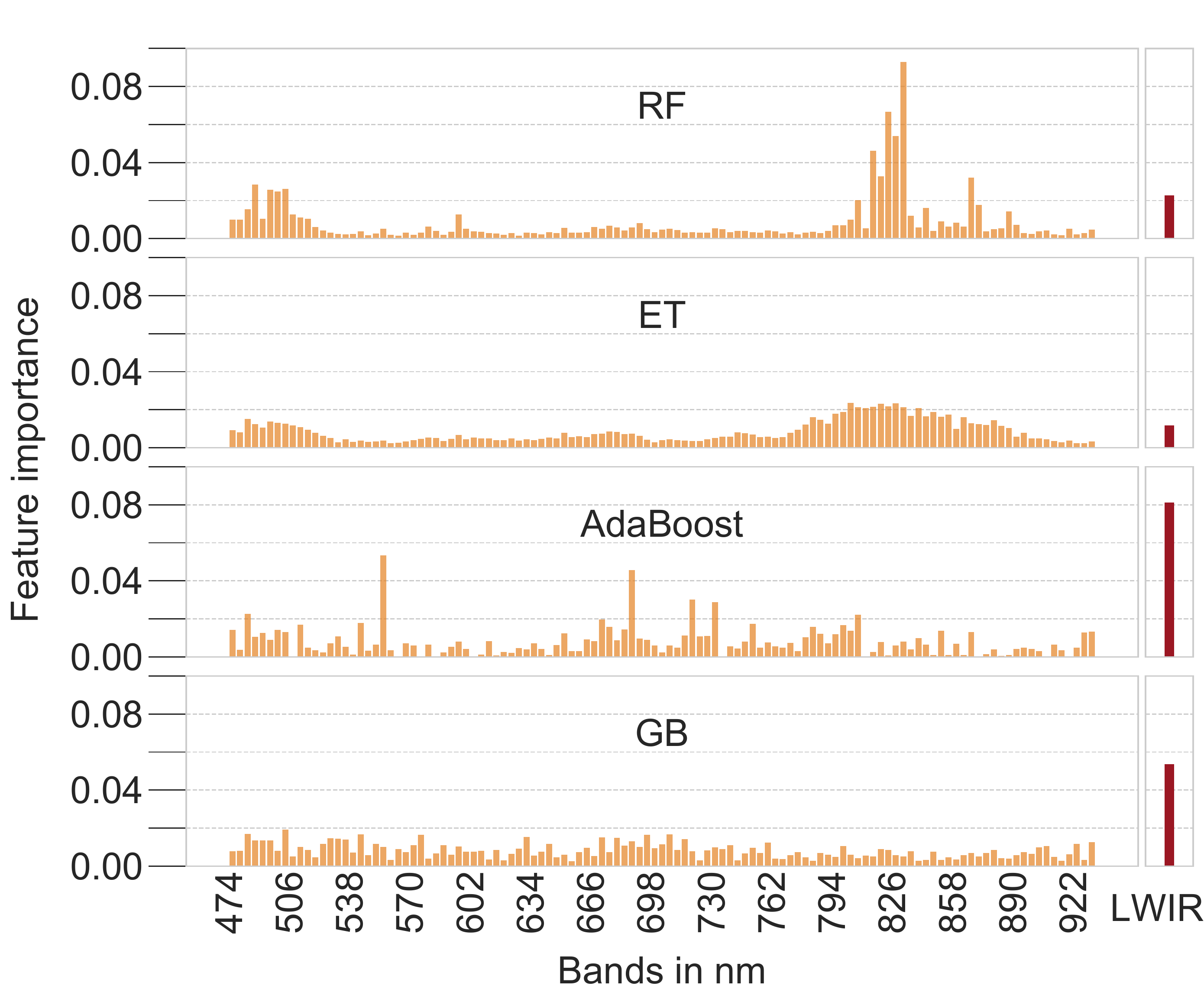}
	\caption{Feature importance of the RF regressor, the ET regressor, the AdaBoost regressor, and the GB regressor. \label{fig:ensemble_fi}}
\end{figure}

Within the ensemble models, RF and ET achieve good regression results. 
ET as an extension of the RF provides the best performance without preprocessing.
An example of the relationship between the estimated and measured soil-moisture values of the ET regressor is given in \Cref{fig:2dhist}.
GB estimates soil moisture slightly better than the AdaBoost.
The influence of min-max scaling on ensemble models is negligible.
\Cref{fig:ensemble_fi} shows the feature importance of the input variables of the baseline framework provided by the ensemble models.
As expected, RF and ET as averaging ensemble models prioritize similar features (input variables).
The distributions of the feature importance of both boosting models AdaBoost and GB differ.
Therefore, the linkages between the spectral features and the underlying physical processes ask for a detailed further analyses.
Strong correlations between the input features (hyperspectral spectral and the LWIR values) appear more challenging for the performance of both boosting models than for the performance of the averaging models.

\begin{figure*}[p]
	\centering
	\includegraphics[width=1.0\textwidth]{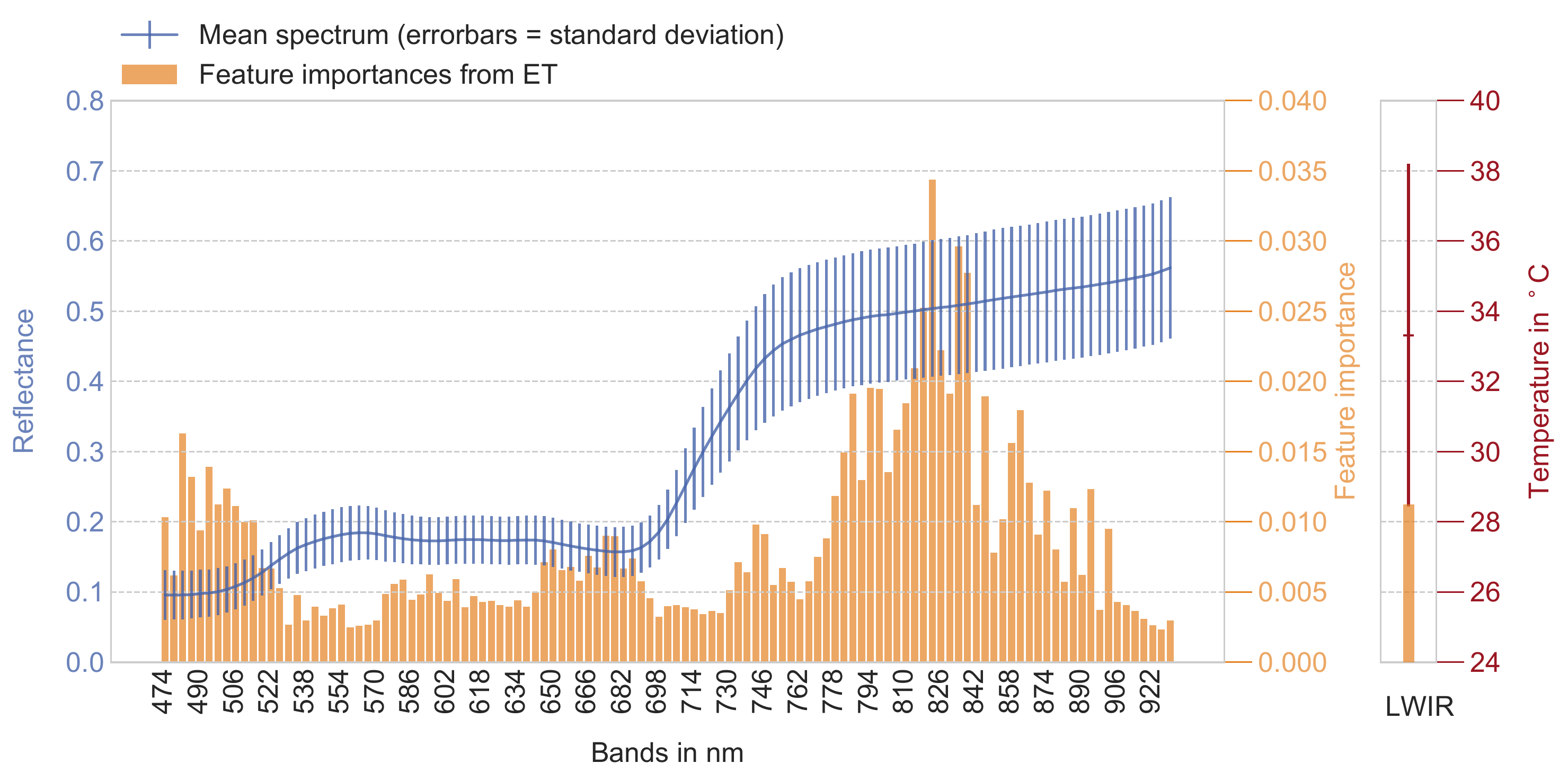}
	\caption{Mean spectrum with the standard deviation as vertical error bars of the hyperspectral data (blue) and the LWIR data (red) of the complete dataset. The feature importance of the ET regressor is shown in orange.\label{fig:featureimportance_et}}
\end{figure*}

In addition, \Cref{fig:featureimportance_et} shows a combination of the hyperspectral mean spectrum with the standard deviation and the feature importance of the ET regressor. 
On the right side, the LWIR data and its importance for the regression is illustrated.
The spectral bands in the area of \SI{826}{\nano\meter} are significantly more important than the remaining ones.
While the variance of the mean spectrum is relatively large in the upper third of the spectrum, the feature importance distribution continually decreases after the identified peak. 
Bands between \SIrange{890}{930}{\nano\meter} possess a minor feature importance and a large variance. 
This finding indicates that these bands exhibit noise, e.g. occurring due to weather and sensor conditions.
The feature importance of the LWIR value plays a minor role.

The k-NN model generally works well and improves with min-max-scaled data due to the linkage between the integrated distance measure and the normalized data.
With respect to the SVM, ANN, and the SOM models, the same effects apply. 

Considering the complete performance of the regression framework, we state that the preprocessing with the PCA-based dimensionality is insufficient of solving the present regression problem. 
The normalization with the min-max scaling seems more favorable for estimating soil moisture based on hyperspectral and LWIR data.
This scaling yields the best regression results for almost any regression model.
We can address an additional preprocessing method by combining a min-max scaling and a PCA as well as other preprocessing techniques and their effects on the regression performance in further studies.
We would like to point out that improving the tuning process of the hyperparameters could further enhance the regression results.

In addition, we remark that the regression framework is data-driven.
It estimates soil-moisture values based on pure reflectance spectra without relying on additional information, e.g. vegetation spectra and information of the measurement angle.
Another notable aspect of estimating soil moisture appears when focussing on the spatial soil-moisture distribution.
This distribution highly depends on further factors such as coverages with mixed vegetation or soil structure.
Thus, it could be extremely inhomogeneous even within small areas.
Furthermore, we would like to take a glance look on the accuracy of the measured soil-moisture reference data. 
According to specification of the installed TDR sensors, their soil-moisture measuring accuracy varies between \num{1}~p.p. to \num{2}~p.p. depending on the moisture values.
Such measurement errors result in a number of effects with respect to the regression framework and finally to the estimation performance.
These effects should be investigated in further work.

\section{Conclusion}
\label{sec:conclusion}

In this paper, we address the estimation of soil moisture based on a measured, pedon-scale dataset.
In contrast to most datasets applied in the context of estimating soil moisture with hyperspectral data, the underlying data consists of VNIR hyperspectral data combined with LWIR data.
The hyperspectral data includes spectral signatures of a vegetated soil surface to ensure an application under real-world conditions prevailing e.g. at a catchment area. 
Our main objective is to investigate the potential of solving the regression problem solely with this data.

We introduce an appropriate regression framework involving one (optional) preprocessing step and nine supervised regression models as well as one model which combines an unsupervised SOM and a supervised SOM. 
The results of the regression framework reveal the potential of respective data-driven models in combination with the used input data under varying real-world measurement circumstances.
In this context, machine learning provides a data-driven solution without exclusively relying on domain-knowledge.

The following challenges are mastered satisfactorily:
\begin{compactitem}
	\item the limited size of data, 
	\item their VNIR spectrum range which is suboptimal referring to preceding studies, 
	\item the fact that we estimate soil moisture with an actively vegetated surface which also is suboptimal to preceding studies, and
	\item the measurement angles differing for each plot. 
\end{compactitem}

To conclude, we point out that it is possible to retrieve soil-moisture content from measured VNIR hyperspectral data  due to the outweigh of the benefits. 

As a direct consequence, we will approach further improvements in different research directions which we point out in the discussion section (cf. \Cref{sec:results}). 
In future work, we plan to analyze in detail the impacts of inhomogeneous soil-moisture distributions and the error propagation which starts with the soil-moisture measuring accuracy of the sensors and relying on this data as reference.
Thereby, we also intend to conduct a dataset on an analog field experiment but using a SWIR sensor for the reflectance measurements. 
Then, we are able to evaluate the performance of the presented regression framework in this dataset and are able to compare the performance with SWIR and VNIR input data.

{\footnotesize
	\begin{spacing}{0.9}
		\bibliography{isprs_final}

\begin{thebibliography}{xx}

\bibitem[Abadi et al., 2016]{abadi2016tensorflow}
Abadi, M., Barham, P., Chen, J., Chen, Z., Davis, A., Dean, J., Devin, M.,
  Ghemawat, S., Irving, G., Isard, M., Kudlur, M., Levenberg, J., Monga, R.,
  Moore, S., Murray, D.~G., Steiner, B., Tucker, P., Vasudevan, V., Warden, P.,
  Wicke, M., Yu, Y. and Zhang, X., 2016.
\newblock Tensorflow: A system for large-scale machine learning.

\bibitem[Altman, 1992]{altman1992an}
Altman, N.~S., 1992.
\newblock An introduction to kernel and nearest-neighbor nonparametric
  regression.
\newblock {\em The American Statistician} 46(3), pp.~175.

\bibitem[Baghdadi et al., 2012]{baghdadi2012estimation}
Baghdadi, N., Cresson, R., Hajj, M.~E., Ludwig, R. and Jeunesse, I.~L., 2012.
\newblock Estimation of soil parameters over bare agriculture areas from c-band
  polarimetric sar data using neural networks.
\newblock {\em Hydrology and Earth System Sciences} 16(6), pp.~1607--1621.

\bibitem[Breiman, 1997]{breiman1997arcing}
Breiman, L., 1997.
\newblock Arcing the edge.
\newblock Technical Report 486, Statistics Department, University of California
  at Berkeley.

\bibitem[Breiman, 2001]{breiman2001random}
Breiman, L., 2001.
\newblock Random forests.
\newblock {\em Machine Learning} 45(1), pp.~5--32.

\bibitem[Brooks et al., 2015]{brooks2015hydrological}
Brooks, P.~D., Chorover, J., Fan, Y., Godsey, S.~E., Maxwell, R.~M., McNamara,
  J.~P. and Tague, C., 2015.
\newblock Hydrological partitioning in the critical zone: Recent advances and
  opportunities for developing transferable understanding of water cycle
  dynamics.
\newblock {\em Water Resources Research} 51(9), pp.~6973--6987.

\bibitem[Cavagnaro, 2016]{cavagnaro2016soil}
Cavagnaro, T.~R., 2016.
\newblock Soil moisture legacy effects: Impacts on soil nutrients, plants and
  mycorrhizal responsiveness.
\newblock {\em Soil Biology and Biochemistry} 95, pp.~173--179.

\bibitem[Colini et al., 2014]{colini2014hyperspectral}
Colini, L., Spinetti, C., Amici, S., Buongiorno, M., Caltabiano, T., Doumaz,
  F., Favalli, M., Giammanco, S., Isola, I., La~Spina, A. et~al., 2014.
\newblock Hyperspectral spaceborne, airborne and ground measurements campaign
  on mt. etna: multi data acquisitions in the frame of prisma mission (asi-agi
  project n. i/016/11/0).
\newblock {\em Quaderni di Geofisica} 119, pp.~1--51.

\bibitem[Dalal and Henry, 1986]{dalal1986simultaneous}
Dalal, R.~C. and Henry, R.~J., 1986.
\newblock Simultaneous determination of moisture, organic carbon, and total
  nitrogen by near infrared reflectance spectrophotometry.
\newblock {\em Soil Science Society of America Journal} 50(1), pp.~120--123.

\bibitem[Fabre et al., 2015]{fabre2015estimation}
Fabre, S., Briottet, X. and Lesaignoux, A., 2015.
\newblock Estimation of soil moisture content from the spectral reflectance of
  bare soils in the $0.4-2.5\upmu\text{m}$ domain.
\newblock {\em Multidisciplinary Digital Publishing Institute} 15(2),
  pp.~3262–--3281.

\bibitem[Farrelly et al., 2011]{farrelly2011sitka}
Farrelly, N., Dhubh\'ain, A.~N. and Nieuwenhuis, M., 2011.
\newblock Sitka spruce site index in response to varying soil moisture and
  nutrients in three different climate regions in ireland.
\newblock {\em Forest Ecology and Management} 262(12), pp.~2199--2206.

\bibitem[Finn et al., 2011]{finn2011remote}
Finn, M.~P., Lewis, M., Bosch, D.~D., Giraldo, M., Yamamoto, K., Sullivan,
  D.~G., Kincaid, R., Luna, R., Allam, G.~K., Kvien, C. and Williams, M.~S.,
  2011.
\newblock Remote sensing of soil moisture using airborne hyperspectral data.
\newblock {\em GIScience \& Remote Sensing} 48(4), pp.~522---540.

\bibitem[Flerchinger and Hardegree, 2004]{flerchinger2004modelling}
Flerchinger, G. and Hardegree, S., 2004.
\newblock Modelling near-surface soil temperature and moisture for germination
  response predictions of post-wildfire seedbeds.
\newblock {\em Journal of Arid Environments} 59(2), pp.~369--385.

\bibitem[Freund and Schapire, 1997]{freund1997a}
Freund, Y. and Schapire, R., 1997.
\newblock A decision-theoretic generalization of on-line learning and an
  application to boosting.
\newblock  55(1), pp.~119--139.

\bibitem[Friedman et al., 2001]{friedman2001elements}
Friedman, J., Hastie, T. and Tibshirani, R., 2001.
\newblock {\em The elements of statistical learning}.
\newblock Vol.~1, Springer series in statistics New York.

\bibitem[Geurts et al., 2006]{geurts2006extremely}
Geurts, P., Ernst, D. and Wehenkel, L., 2006.
\newblock Extremely randomized trees.
\newblock {\em Machine Learning} 63(1), pp.~3--42.

\bibitem[Gill et al., 2006]{gill2006soil}
Gill, M.~K., Asefa, T., Kemblowski, M.~W. and McKee, M., 2006.
\newblock Soil moisture prediction using support vector machines.
\newblock {\em JAWRA Journal of the American Water Resources Association}
  42(4), pp.~1033--1046.

\bibitem[Girardin and Wotton, 2009]{girardin2009summer}
Girardin, M.~P. and Wotton, B.~M., 2009.
\newblock Summer moisture and wildfire risks across canada.
\newblock {\em Journal of Applied Meteorology and Climatology} 48(3),
  pp.~517--533.

\bibitem[Guanter et al., 2015]{guanter2015the}
Guanter, L., Kaufmann, H., Segl, K., Foerster, S., Rogass, C., Chabrillat, S.,
  Kuester, T., Hollstein, A., Rossner, G., Chlebek, C., Straif, C., Fischer,
  S., Schrader, S., Storch, T., Heiden, U., Mueller, A., Bachmann, M., Mühle,
  H., Müller, R., Habermeyer, M., Ohndorf, A., Hill, J., Buddenbaum, H.,
  Hostert, P., Linden, S. v.~d., Leitão, P.~J., Rabe, A., Doerffer, R.,
  Krasemann, H., Xi, H., Mauser, W., Hank, T., Locherer, M., Rast, M., Staenz,
  K. and Sang, B., 2015.
\newblock {The EnMAP Spaceborne Imaging Spectroscopy Mission for Earth
  Observation}.
\newblock {\em Remote Sensing} 7(7), pp.~8830--8857.

\bibitem[Haubrock, 2008]{haubrock2008surface}
Haubrock, S.-N., 2008.
\newblock Surface soil moisture quantification and validation based on
  hyperspectral data and field measurements.
\newblock {\em Journal of Applied Remote Sensing} 2(1), pp.~1--26.

\bibitem[Ifarraguerri and Chang, 2000]{ifarraguerri2000unsupervised}
Ifarraguerri, A. and Chang, C.-I., 2000.
\newblock Unsupervised hyperspectral image analysis with projection pursuit.
\newblock {\em IEEE Transactions on Geoscience and Remote Sensing} 38(6),
  pp.~2529--2538.

\bibitem[Jackisch et al., 2017]{jackisch2017form}
Jackisch, C., Angermann, L., Allroggen, N., Sprenger, M., Blume, T., Tronicke,
  J. and Zehe, E., 2017.
\newblock Form and function in hillslope hydrology: in situ imaging and
  characterization of flow-relevant structures.
\newblock {\em Hydrology and Earth System Sciences} 21(7), pp.~3749--3775.

\bibitem[Jarvis, 2007]{jarvis2007a}
Jarvis, N.~J., 2007.
\newblock A review of non‐equilibrium water flow and solute transport in soil
  macropores: principles, controlling factors and consequences for water
  quality.
\newblock {\em European Journal of Soil Science} 58(3), pp.~523--546.

\bibitem[John, 1992]{john1992soil}
John, B., 1992.
\newblock Soil moisture detection with airborne passive and active microwave
  sensors.
\newblock {\em International Journal of Remote Sensing} 13(3), pp.~481--491.

\bibitem[Kaleita et al., 2005]{kaleita2005relationship}
Kaleita, A.~L., Tian, L.~F. and Hirschi, M.~C., 2005.
\newblock Relationship between soil moisture content and soil surface
  reflectance.
\newblock {\em Transactions of the ASAE} 48(5), pp.~1979--1986.

\bibitem[Keller et al., 2018]{keller2018modeling}
Keller, S., Riese, F.~M., Allroggen, N., Jackisch, C. and Hinz, S., 2018.
\newblock Modeling subsurface soil moisture based on hyperspectral data: First
  results of a multilateral field campaign.
\newblock In: \emph{Tagungsband der 37. Wissenschaftlich-Technische
  Jahrestagung der DGPF e.V.}, Vol.~27, pp.~34--48.

\bibitem[Kohonen, 1990]{kohonen1990the}
Kohonen, T., 1990.
\newblock The self-organizing map.
\newblock  78(9), pp.~1464–--1480.

\bibitem[Maggioni et al., 2006]{maggioni2006multi}
Maggioni, V., Panciera, R., Walker, J.~P., Rinaldi, M., Paruscio, V., Kalma,
  J.~D., Kim, E.~J. et~al., 2006.
\newblock A multi-sensor approach for high resolution airborne soil moisture
  mapping.
\newblock In: \emph{30th Hydrology \& Water Resources Symposium: Past, Present
  \& Future}, Conference Design, pp.~297--302.

\bibitem[Massari et al., 2014]{massari2014potential}
Massari, C., Brocca, L., Moramarco, T., Tramblay, Y. and Lescot, J.-F.~D.,
  2014.
\newblock Potential of soil moisture observations in flood modelling:
  Estimating initial conditions and correcting rainfall.
\newblock {\em Advances in Water Resources} 74, pp.~44--53.

\bibitem[Oltra-Carrió et al., 2015]{oltracarri2015improvement}
Oltra-Carrió, R., Baup, F., Fabre, S., Fieuzal, R. and Briottet, X., 2015.
\newblock Improvement of soil moisture retrieval from hyperspectral vnir-swir
  data using clay content information: From laboratory to field experiments.
\newblock {\em Remote Sensing} 7(3), pp.~3184--3205.

\bibitem[Pasolli et al., 2014]{pasolli2014polarimetric}
Pasolli, L., Notarnicola, C., Bruzzone, L., Bertoldi, G., Chiesa, S.~D.,
  Niedrist, G., Tappeiner, U. and Zebisch, M., 2014.
\newblock Polarimetric radarsat-2 imagery for soil moisture retrieval in alpine
  areas.
\newblock {\em Canadian Journal of Remote Sensing} 37(5), pp.~535--547.

\bibitem[Pedregosa et al., 2011]{pedregosa2011scikitlearn}
Pedregosa, F., Varoquaux, G., Gramfort, A., Michel, V., Thirion, B., Grisel,
  O., Blondel, M., Louppe, G., Prettenhofer, P., Weiss, R., Dubourg, V.,
  Vanderplas, J., Passos, A., Cournapeau, D., Brucher, M., Perrot, M. and
  Duchesnay, E., 2011.
\newblock Scikit-learn: Machine learning in python.
\newblock {\em Journal of Machine Learning Research} 12, pp.~2825--2830.

\bibitem[Riese and Keller, 2018]{riese2018introducing}
Riese, F.~M. and Keller, S., 2018.
\newblock Introducing a framework of self-organizing maps for regression of
  soil moisture with hyperspectral data.
\newblock In: \emph{2018 IEEE International Geoscience and Remote Sensing
  Symposium (IGARSS)}.
\newblock Accepted.

\bibitem[Robinson et al., 2008]{robinson2008soil}
Robinson, D.~A., Campbell, C.~S., Hopmans, J.~W., Hornbuckle, B.~K., Jones,
  S.~B., Knight, R., Ogden, F., Selker, J. and Wendroth, O., 2008.
\newblock Soil moisture measurement for ecological and hydrological
  watershed-scale observatories: A review.
\newblock {\em Vadose Zone Journal} 7(1), pp.~358--389.

\bibitem[Salisbury and D'Aria, 1992]{salisbury1992emissivity}
Salisbury, J.~W. and D'Aria, D.~M., 1992.
\newblock Emissivity of terrestrial materials in the $8–14 \mu$m atmospheric
  window.
\newblock {\em Remote Sensing of Environment} 42(2), pp.~83--106.

\bibitem[Tian et al., 2018]{tian2018soil}
Tian, L., Zhao, L., Wu, X., Fang, H., Zhao, Y., Hu, G., Yue, G., Sheng, Y., Wu,
  J., Chen, J., Wang, Z., Li, W., Zou, D., Ping, C.-L., Shang, W., Zhao, Y. and
  Zhang, G., 2018.
\newblock Soil moisture and texture primarily control the soil nutrient
  stoichiometry across the tibetan grassland.
\newblock {\em Science of The Total Environment} 622, pp.~192--202.

\bibitem[Vapnik, 1995]{vapnik1995the}
Vapnik, V.~N., 1995.
\newblock {\em The Nature of Statistical Learning Theory}.
\newblock Springer-Verlag New York, Inc., New York, NY, USA.

\bibitem[Vereecken et al., 2014]{vereecken2014on}
Vereecken, H., Huisman, J., Pachepsky, Y., Montzka, C., Kruk, J. v.~d., Bogena,
  H., Weihermüller, L., Herbst, M., Martinez, G. and Vanderborght, J., 2014.
\newblock On the spatio-temporal dynamics of soil moisture at the field scale.
\newblock {\em Journal of Hydrology} 516, pp.~76--96.

\bibitem[Wang et al., 2007]{wang2007a}
Wang, L., Jia, X. and Zhang, Y., 2007.
\newblock A novel geometry-based feature-selection technique for hyperspectral
  imagery.
\newblock {\em IEEE Geoscience and Remote Sensing Letters} 4(1), pp.~171--175.

\bibitem[Wang et al., 2014]{wang2014freezethaw}
Wang, L., Shi, Z., Wu, G. and Fang, N., 2014.
\newblock Freeze/thaw and soil moisture effects on wind erosion.
\newblock {\em Geomorphology} 207, pp.~141--148.

\bibitem[Xie et al., 2014]{xie2014soil}
Xie, X.~M., Xu, J.~W., Zhao, J.~F., Liu, S. and Wang, P., 2014.
\newblock Soil moisture inversion using amsr-e remote sensing data: An
  artificial neural network approach.
\newblock {\em Applied Mechanics and Materials} 501-504, pp.~2073--2076.

\end{thebibliography}
	\end{spacing}
}

\section*{APPENDIX} 
\label{sec:app}

The appendix contains the setup of the hyperparameters for all regression models without preprocessing, cf. \Cref{tab:hyperparams}.

\renewcommand{\arraystretch}{1.5}
\begin{table*}[bp]
	\centering
	\caption{Hyperparameter setup for the regression framework without preprocessing. This setup is obtained by a basic grid search algorithm with 10-fold cross validation on the training subset. The regressors are implemented mostly in scikit-learn~\citep{pedregosa2011scikitlearn} and TensorFlow~\citep{abadi2016tensorflow}, while the SOM is implemented according to~\citet{riese2018introducing}.}
	\begin{tabular}{llcl}
		\toprule
		Model & Reference & Package & Hyperparameter setup\\
		\midrule
		Linear 		& -- 						& scikit-learn &  --\\    
		PLS 		& --						& scikit-learn & $\texttt{n\_components} = 10;\ \texttt{max\_iter} = 100;\ \texttt{tol} = 10^{-7}$\\     
		RF 			& \cite{breiman2001random} 	& scikit-learn & $\texttt{n\_estimators} = 1000$\\
		ET 			& \cite{geurts2006extremely}& scikit-learn & $\texttt{n\_estimators} = 1000$\\
		AdaBoost 	& \cite{freund1997a}		& scikit-learn & $\texttt{learning\_rate} = 3.0; \texttt{loss} = \text{"linear"}; \texttt{n\_estimators} =150$\\
		GB 			& \cite{breiman1997arcing}	& scikit-learn & \begin{minipage}[t]{0.5\linewidth} $\texttt{learning\_rate} = 0.1;\ \texttt{loss} = \text{"huber"};$ \\ $\texttt{n\_estimators} = 1000;\ \texttt{max\_depth} = 2$  \end{minipage}\\
		k-NN 		& \cite{altman1992an}		& scikit-learn & $\texttt{n\_neighbors} = 6;\ \texttt{weights} = \text{"distance"};\ \texttt{leaf\_size} = 1$\\
		SVM 		& \cite{vapnik1995the}		& scikit-learn & $C=26827;\ \gamma=0.00178$\\
		ANN 		& \cite{friedman2001elements} & TensorFlow & \begin{minipage}[t]{0.5\linewidth} Keras sequential model with $\texttt{epochs} = 70;\ \texttt{batch\_size} = 8$; four dense layers with $\{64, 128, 64, 32\}$ neurons and RELU activations\end{minipage}\\
		SOM 		&  \begin{minipage}[t]{0.18\linewidth}\citet{kohonen1990the};\\ \citet{riese2018introducing} \end{minipage}& other & \begin{minipage}[t]{0.5\linewidth}SOM size $= 30\times70$;\ $N_{\text{Iterations, Input}} = 5000$;\ $N_{\text{Iterations, Output}} = 8000$;\ learning rates $\alpha_{\text{Start}}=0.4;\  \alpha_{\text{End}} = 0.005$;\ exponential neighborhood function (input and output);\ pseudo-gaussian neighborhood distance weight \end{minipage}\\
		\bottomrule
	\end{tabular}
	\label{tab:hyperparams}
\end{table*}

\end{document}